\documentclass{article}

 \usepackage[sglblindworkshop, final]{neurips_2025}
\usepackage[utf8]{inputenc} % allow utf-8 input
\usepackage[T1]{fontenc}    % use 8-bit T1 fonts
\usepackage{hyperref}       % hyperlinks
\usepackage{url}            % simple URL typesetting
\usepackage{booktabs}       % professional-quality tables
\usepackage{amsfonts}       % blackboard math symbols
\usepackage{nicefrac}       % compact symbols for 1/2, etc.
\usepackage{microtype}      % microtypography

\usepackage{graphicx}
\usepackage{amsmath}
\usepackage{float}

\usepackage[dvipsnames]{xcolor} 
\usepackage{tabularx}
\usepackage{tcolorbox}
\tcbuselibrary{listings,breakable}
\usepackage{multirow}

\usepackage{enumitem}
\title{PepThink-R1: LLM for Interpretable Cyclic Peptide Optimization with CoT SFT and Reinforcement Learning\thanks{Copyright © 2025 Merck \& Co., Inc., Rahway, NJ, USA and its affiliates.}
\thanks{PepThink-R1 code is available at \url{https://github.com/MSDLLCpapers/PepThink-R1}}}
\workshoptitle{AI for Science}

\author{%
    Ruheng Wang\textsuperscript{\rm 1,2}
    Hang Zhang\textsuperscript{\rm 1,3}  
    Trieu Nguyen\textsuperscript{\rm 1} 
     \\
    \textbf{
    Shasha Feng\textsuperscript{\rm 1} 
    Hao-Wei Pang\textsuperscript{\rm 1} 
    Xiang Yu\textsuperscript{\rm 1}
    Li Xiao\textsuperscript{\rm 1}
    Peter Zhiping Zhang\textsuperscript{\rm 1*}}\\
    \textsuperscript{\rm 1}Merck \& Co., Inc., Rahway, NJ, USA\\
    \textsuperscript{\rm 2}UT Southwestern Medical Center, Dallas, TX, USA\\
    \textsuperscript{\rm 3}University of Pittsburgh, Pittsburgh, PA, USA\\
    \texttt{ruheng.wang@utsouthwestern.edu, zhiping.peter.zhang@merck.com}
}

\begin{document}

\maketitle

\begin{abstract}
Designing therapeutic peptides with tailored properties is hindered by the vastness of sequence space, limited experimental data, and poor interpretability of current generative models. To address these challenges, we introduce \textbf{PepThink-R1}, a generative framework that integrates large language models (LLMs) with chain-of-thought (CoT) supervised fine-tuning and reinforcement learning (RL). Unlike prior approaches, PepThink-R1 explicitly reasons about monomer-level modifications during sequence generation, enabling interpretable design choices while optimizing for multiple pharmacological properties. Guided by a tailored reward function balancing chemical validity and property improvements, the model autonomously explores diverse sequence variants. We demonstrate that PepThink-R1 generates cyclic peptides with significantly enhanced lipophilicity, stability, and exposure, outperforming existing general LLMs (e.g., GPT-5) and domain-specific baseline in both optimization success and interpretability. To our knowledge, this is the first LLM-based peptide design framework that combines explicit reasoning with RL-driven property control, marking a step toward reliable and transparent peptide optimization for therapeutic discovery.

\end{abstract}

%-------------
%%%Introduction %%%
%-------------

\section{Introduction}

Peptide-based therapeutics are gaining momentum in modern drug discovery due to their high specificity, tunable bioactivity, and ability to target challenging protein–protein interactions\cite{wang2022therapeutic}. 
Cyclic peptides, in particular, offer enhanced stability and binding affinity from their constrained conformations and the incorporation of non-natural amino acids (NNAAs)\cite{ji2024cyclic}. 
Despite this promise, designing peptides with multiple desired properties, such as solubility and binding affinity, remains a formidable challenge due to the vastness and complexity of the design space and the scarcity of labeled data for optimization\cite{muttenthaler2021trends}.

The application of artificial intelligence (AI), especially generative models, has shown remarkable progress in addressing these challenges\cite{lin2025highplay, gong2025pepseek, wang2025discovery, xu2024helm}. 
Tools such as PepTune\cite{tang2025peptune} and PepINVENT\cite{geylan2025pepinvent} represent two notable advances in the field. 
PepTune leverages discrete masked diffusion models and Monte Carlo Tree Search (MCTS) for multi-objective optimization of peptide SMILES strings, enabling the design of chemically modified and cyclic peptides in a discrete representation space. 
Similarly, PepINVENT extends the REINVENT\cite{olivecrona2017molecular,loeffler2024reinvent} platform to peptides with a masked language model as an idea generator, enabling peptide optimization at specified monomer position. 
While the impact of these tools on real-world projects remains to be seen, a persistent limitation of current peptide design frameworks is their lack of interpretable reasoning during generation.
Without insights into why a particular sequence of modifications enhances a given design objective, such models can be difficult to trust or adapt to program-specific constraints. 

Inspired by recent advancement in reasoning in Large Language Models (LLMs)\cite{guo2025deepseek, wei2022chain}, we propose a novel framework that integrates LLM with explicit chain-of-thought (CoT) supervised fine-tuning (SFT) and reinforcement learning (RL) to improve interpretable property controlled cyclic peptide optimization. 
Unlike previous work, our approach directly leverages monomer-level reasoning during sequence generation, offering improved interpretability and modularity, which is critical for iterative design in real-world therapeutic applications. Specifically, our contributions include: 

\begin{enumerate}
    \item A data construction pipeline that transforms raw peptide data into reasoning-augmented peptide pairs, enabling SFT with property-centric monomer modification rationales.
    \item A simple yet effective CoT construction strategy that aligns monomer-level modifications with desired property improvements, providing a semantic scaffold for guided generation.
    \item Supervised fine-tuning of an LLM on the constructed dataset, enabling the model to internalize chemically meaningful reasoning patterns.
    \item A reinforcement learning module that encourages the LLM to apply and refine its reasoning when exploring novel peptide sequences, guided by a tailored reward function combining chemical validity and multi-property optimization.
\end{enumerate}

Our method represents, to the best of our knowledge, the first LLM-based peptide generator that explicitly integrates CoT reasoning and RL for monomer-level optimization. By combining symbolic reasoning with sequence exploration, we improved interpretable generation of cyclic peptides with desired properties. 
Preliminary results demonstrate substantial improvements in property satisfaction rates over state-of-the-art methods, marking a promising step toward accelerating peptide design cycle.

%-------------
%%%Related work %%%
%-------------
\section{Related Work}
Designing functional peptides, including cyclic peptides, is challenging due to the immense sequence space and complex property requirements. Recent work has explored large language models (LLMs), variational autoencoders, and other deep generative models as the generator for new peptides, and using techniques such as RL\cite{geylan2025pepinvent}, MCTS\cite{tang2025peptune, lin2025highplay}, genetic algorithms~\cite{krishnan2025generative} and LLM\cite{gong2025pepseek} to guide exploration of peptide sequence space toward desired properties. 

\paragraph{Peptide generation.}

The majority of work represents peptide as linear sequence in HELM\cite{zhang2012helm} or SMILES\cite{weininger1988smiles}, and apply generative modeling to design peptides with tailored therapeutic properties. 
Several models, such as HELM-GPT\cite{xu2024helm} and AMP-Designer\cite{wang2025discovery}, are GPT-based models that operate at monomer level using HELM representation. 
SMILES-based representation is adopted by PepINVENT and PepTune, both leverage masked language models for diverse peptide generation. 
Some work utilizes the conversion between different representation forms to enable generation in sequence while evaluation with three-dimensional structural information\cite{lin2025highplay, chen2025target}.
More related work can be found in reviews by Wan, Zhai, et al.~\cite{wan2022deep, zhai2025artificial, wan2024machine}.

\paragraph{RL for multi-property optimization in chemical space.}
Recent work\cite{loeffler2024reinvent, wang2024reinforcement, wang2023self, popova2018deep} has shown that reinforcement learning (RL) is an effective strategy for multi-property optimization in therapeutic entity design, enabling goal-directed discovery in multiple modalities including small molecule, peptide, and protein. 
Many works\cite{wang2025discovery, tang2025peptune, geylan2025pepinvent} integrate RL modules to fine-tune generative models with feedback from property predictors to optimize traits like cell permeability, antimicrobial activity, and binding affinity.
Techniques range from policy gradient and reward shaping to inference-time search strategies like Monte Carlo Tree Guidance\cite{tang2025peptune}, allowing models to balance trade-offs across multiple, often conflicting therapeutic objectives.

\paragraph{Chain-of-Thought SFT for chemical generation.}
Chain-of-Thought (CoT)\cite{wei2022chain} aims to enhance the reasoning capability of LLMs by generating intermediate reasoning steps before arriving at a final answer and is now a key technology in enhancing reasoning capability in modern LLMs\cite{guo2025deepseek, yeo2025demystifying}. 
A few recent works have extended CoT reasoning to address biochemistry related problems, including chemical question-answering tasks\cite{ouyang2023structured}, protein-protein interaction prediction\cite{jin2024prollm}, molecular structural understanding\cite{jang2024chain}, property prediction\cite{jin2025effective}.
Directly incorporating CoT into SFT of LLMs for chemical optimization remains unseen to the best of our knowledge.

%-------------
%%%mMethodology%%%
%-------------
\section{Methodology}

\begin{figure}[!htbp]
  \centering
  \includegraphics[trim=0cm 2cm 0cm 0.8cm,clip,width=\linewidth]{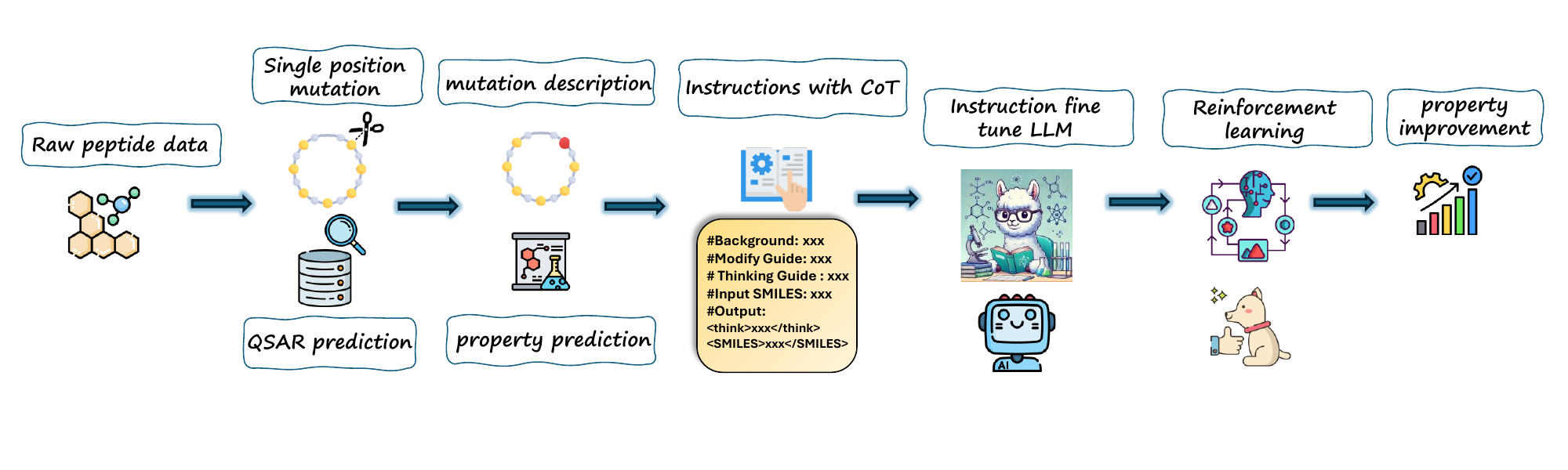}
  \caption{\textbf{Overview of our approach.}
  Peptide pairs were constructed from single position mutation of raw data with property values predicted by QSAR model. Those pairs were used for CoT SFT of pretrained LLM, which was further improved by RL and used for peptide optimization.}
  \label{fig:method}
\end{figure}

The overview of our approach is depicted in Figure.\ref{fig:method}. 
We first augmented the raw peptide data from CycPeptMPDB database~\cite{li2023cycpeptmpdb} through single position random mutation of peptide HELM~\cite{zhang2012helm} code to generate peptide pairs with single monomer difference, and then used internal QSAR model~\cite{heid2023chemprop} to get property values of all peptides.
High quality peptide pairs were used for constructing training data for CoT SFT of pretrained LLM~\cite{yu2024llasmol}. To further improve property control of generated peptides, an RL strategy was designed to further optimize the model. 
Details of our training data construction strategy, design of CoT prompt, and RL approach are introduced below.

%%%%%%%%%%
\subsection{Preparing synthetic training dataset}
\label{subsec:gen_train_data}

\paragraph{Raw data.}
We constructed the dataset starting from 5,530 head-to-tail cyclic peptide sequences extracted from the CycPeptMPDB database~\cite{li2023cycpeptmpdb}. 
After removing sequences that can not be processed by CycloPs~\cite{duffy2011cyclops}, 3,778 peptides composed solely of natural $\alpha$-amino-acid (AA) monomers were retained.
To expand the chemical space, we incorporated non-natural $\alpha$-amino acids (NNAAs) from the virtual library developed by Amarasinghe et al.~\cite{amarasinghe2022virtual}. This reaction-based library contains approximately 380,000 synthesizable NNAAs, from which we used a representative subset of 10,000. Combined with 385 natural monomers from CycPeptMPDB, this resulted in a building block vocabulary of 10,385 unique monomers.

\paragraph{Data augmentation.}
Peptide augmentation was carried out using CycloPs~\cite{duffy2011cyclops}. For each of the 3,778 natural peptides, we applied 100 independent point mutations at randomly selected non-terminal positions in their HELM sequences~\cite{zhang2012helm}. In each mutation, the original residue was replaced by a randomly sampled monomer from the monomer vocabulary. 
For every resulting sequence, we recorded the mutated peptide, the mutation index, and the pre- and post-mutation monomer identities. This procedure yielded a raw dataset of approximately 380,000 unique cyclic peptides, and each mutated peptide and its original counterpart naturally became a peptide pair with only single monomer difference. 

\paragraph{Property annotation.} 
All peptides were annotated using a multi-task message-passing QSAR model built upon Chemprop~\cite{heid2023chemprop}, trained for ADMET-relevant property prediction on internal data as reported by Chen et al.~\cite{chen2025data}. We focused on three endpoints predicted by QSAR: (1) \textbf{LogD} ({LogD\_HPLC\_pH\_7}) representing lipophilicity at pH 7; (2) \textbf{MRT} ({Rat\_MRT(hrs)}) representing mean residence time in rats, reported in hours; and (3) \textbf{SIF} ({SIF\_halflife\_10mg\_per\_ml(hrs)}) representing the peptide half-life in simulated intestinal fluid.

\paragraph{Data categorization.}  
To better reflect practical objectives in drug discovery, we discretized each predicted property into three categorical levels: {low}, {medium}, and {high}. The thresholds were empirically chosen as follows based on data distribution:  
\text{LogD} — low: <3, medium: 3–4.2, high: >4.2;  
\text{MRT} — low: <0.56, medium: 0.56–1.63, high: >1.63;  
\text{SIF} — low: <3.4, medium: 3.4–10.1, high: >10.1.
Each peptide pair was categorized based on whether one or more properties improved to the {high} level. For instance, a peptide pair showing improvement of LogD and SIF to high-level while MRT remained unchanged would be labeled as a \textit{LogD up, SIF up} case. We grouped samples into single-, dual-, and triple-property improvement categories.
Compared to raw numerical changes, categorical improvements offer a more practically meaningful evaluation of chemical optimization. In real-world lead optimization, hitting specific ADMET thresholds is often more relevant than small numeric gains.

\paragraph{Training and evaluation datasets.}
To construct the supervised fine-tuning (SFT) dataset, we retained all peptides with triple-property improvement, as they exemplify strong multi-objective success. To enrich the training dataset, we further sampled up to 4,000 peptides from selected dual- and single-property improvement groups, enabling the model to learn diverse patterns of beneficial sequence changes. For RL, we randomly selected a subset of 600 peptides from the SFT training set to serve as the initial training pool.
For evaluation, we excluded all triple-property improvement cases from the test set, resulting in 1,880 hold-out samples derived from lower-performing peptides. This setting allowed us to assess whether the fine-tuned model can generalize enhancement strategies beyond the high-performing examples seen during training.

%%%%%%%%%%%%%%
\subsection{CoT prompt design}
\label{subsec:CoT}

\newcommand{\think}[1]{\textcolor{blue}{\texttt{<think>}} #1 \textcolor{blue}{\texttt{</think>}}}
\newcommand{\smiles}[1]{\textcolor{ForestGreen}{\texttt{<SMILES>}} #1 \textcolor{ForestGreen}{\texttt{</SMILES>}}}

\begin{table}[!htbp]
    \centering
    \renewcommand{\arraystretch}{1.12} 
    \begin{tabular}{p{13cm}}
        \hline
        \textbf{CoT training prompt} \\ \hline

        \textbf{Background:} We are modifying peptides to meet specific ADMET property improvements by reasoning step-by-step.  
        The required answer format is wrapped only within \textcolor{ForestGreen}{\texttt{<SMILES>}} and \textcolor{ForestGreen}{\texttt{</SMILES>}} HTML tags. \\[6pt]

        \textbf{Peptide Modify Guides:} Increase lipophilicity (LogD), mean residence time (MRT\_Rat), and SIF stability. \\[6pt]

        \textbf{Thinking Process Guides:} Please enclose your step-by-step reasoning process within \textcolor{blue}{\texttt{<think>}} and \textcolor{blue}{\texttt{</think>}} HTML tags.  
        This process should clearly explain how you modify the monomer in the original SMILES before providing the final SMILES. \\[6pt]

        \textbf{Input SMILES:} \\
        \smiles{\textit{[the similes of the input peptide]}} 
        \\[8pt]

        \textbf{Output example:} \\ 
        \think{At position \textit{[X]}, the monomer changed from \textit{[smiles of leaving monomer]}
         to \textit{[smiles of new monomer]} to \textit{[description of property changes, e.g., increase lipophilicity (LogD), mean residence time (MRT\_Rat), and SIF stability]}.}\\[4pt]
        \smiles{\textit{[the similes of the output peptide]}} \\ \hline
    \end{tabular}
    \caption{\textbf{Template for constructing PepThink-R1 CoT prompt.} Contents within \textit{[\;]} are populated by each specific peptide pair.}
    \label{tab:cot_prompt}
\end{table}

To train \text{PepThink-R1}, we designed a prompt template that explicitly guides the language model to iteratively improve peptides by reasoning about monomer-level modifications. As shown in Table~\ref{tab:cot_prompt}, this template consists of three main components: a task background, a set of modification objectives, and a detailed reasoning guideline. The model was required to reason step-by-step about how to alter a specific monomer in the peptide sequence, and then output a modified SMILES string wrapped in HTML tags.
Our design encourages the model to treat peptides as modular sequences composed of individual monomers. Instead of allowing unconstrained mutations over the entire SMILES, the prompt limits the scope of changes to monomer substitutions. This prevents overly drastic or chemically implausible edits, while also promoting more interpretable and localized reasoning. Furthermore, this setup encourages the reinforcement learning agent to explore novel monomer designs in a structured way, fostering creativity without sacrificing molecular integrity.

%%%%%%%%%%%%%%%%
\subsection{Reinforcement learning with pharmacology-aware rewards}
\label{subsec:reward_model}

\paragraph{Algorithm.}
For the RL training phase, We fine-tuned our model $\pi_\theta$ using Group Relative Policy Optimization (GRPO) \cite{shao2024deepseekmathpushinglimitsmathematical} to enhance the reasoning process. Following the CoT prompt template, each training sample $x \in \mathcal{D}$ ($\mathcal{D}$ represents training dataset) consists of a SMILES string to be edited $x_s$ and a textual prompt $x_t$. Given input $x = (x_s, x_t)$, the model samples a group of candidate outputs $\{o^{(1)}, \ldots, o^{(G)}\} \sim \pi_\theta(y \mid x)$, where each $o^{(i)}$ includes both reasoning trace and final answer. We define a customized reward function to score each output. Relative scores $\{R_i\}$ are then computed and normalized to obtain advantages $\{A_i\}$, which are used to update $\pi_\theta$ following the GRPO objective (Eq.~\eqref{eq:grpo}) as follows:

{\small
\begin{equation}
\begin{aligned}
\mathcal{J}_{\text{GRPO}}(\theta) = \mathbb{E}_{x\sim{D}, \{y_i\}_{i=1}^{G} \sim \pi_{\text{old}}(y \mid x)} \Bigg[
\frac{1}{G} \sum_{i=1}^{G} \Bigg(
\min\Big(
r_i A_i,\,
\text{clip}(r_i, 1 - \epsilon, 1 + \epsilon) A_i
\Big)
- \beta \, \mathbb{D}_{\text{KL}}(\pi_\theta \| \pi_{\text{ref}})
\Bigg)
\Bigg],
\end{aligned}
\label{eq:grpo}
\end{equation}
}
where \( r_i = \frac{\pi_\theta(y_i \mid x)}{\pi_{\text{old}}(y_i \mid x)}, A_i = \frac{R_i - \text{mean}(\{R_i\}_{i=1}^{G})}{\text{std}(\{R_i\}_{i=1}^{G})} \), \( \epsilon \), \( \beta \) are hyperparameters, and \( \pi_{\text{old}} \) is the policy model used for rollouts sampling.

\paragraph{Reward Modeling.} 
To guide reinforcement learning toward the generation of pharmacologically promising molecules, we designed a reward function that balances \text{property desirability}, \text{structural similarity}, and \text{diversity control} by adapted partially from REINVENT\cite{loeffler2024reinvent}. The final reward $R$ for a candidate peptide is computed as:

\begin{equation}
R = \text{dup}_\text{fac} \cdot \left(0.8 \cdot \text{prop}_\text{smooth} + 0.2 \cdot \text{sim}_\text{fac} \right)
\end{equation}
\noindent where each component is defined as follows.

\vspace{4pt}
\noindent\textit{Property desirability.}  
We employed a Chemprop-based multi-task QSAR model\cite{heid2023chemprop} to predict three key ADMET properties: LogD, MRT\textsubscript{Rat}, and SIF $t_{1/2}$. Instead of using raw regression outputs, we applied a smooth sigmoid-based transformation to capture whether a prediction exceeds a predefined threshold:

\begin{equation}
\text{prop}_\text{smooth} = \frac{1}{3} \sum_{i=1}^{3} \sigma\left( \frac{x_i - t_i}{k_i} \right)
\end{equation}

\noindent where $x_i$ is the predicted value for property $i$, $t_i$ is its threshold (e.g., LogD > 4.2), $k_i$ is a scale factor controlling the sharpness of transition, and $\sigma(\cdot)$ is the logistic function. This component encourages the generation of molecules that simultaneously approach all three pharmacokinetic targets.

\vspace{4pt}
\noindent\textit{Similarity factor.}  
To retain structural resemblance with the input molecule, we computed the Tanimoto similarity $s$ between the Morgan fingerprints of the original and generated peptides. The similarity reward is derived via:

\begin{equation}
\text{sim}_\text{fac} = \sigma\left( \alpha \cdot (s - s_0) \right)
\end{equation}

\noindent where $\alpha$ controls the steepness of the reward curve and $s_0$ is the target similarity center (set to 0.6). This promotes local exploration in chemical space.

\vspace{4pt}
\noindent\textit{Duplication penalty.}  
To prevent mode collapse and encourage structural diversity, we penalized previously generated molecules via a frequency-based factor:

\begin{equation}
\text{dup}_\text{fac} = \left( \frac{1}{\max(1, n + 1)} \right)^{\gamma}
\end{equation}

\noindent where $n$ is the number of times the current molecule has appeared in the generation history (tracked via a least-recently-used cache), and $\gamma$ is a decay exponent.

\vspace{4pt}
\noindent Overall, this reward formulation ensures that generated peptides are not only pharmacologically improved but also remain structurally meaningful and diverse.

\subsection{Evaluation metrics.}

We evaluated the quality of generated peptides using six metrics that reflect chemical validity, diversity, novelty, and optimization performance on three endpoints. 
These metrics together provide a comprehensive assessment of the model's ability to produce valid, novel, diverse, and high-performing peptide molecules in a controlled manner.
Each metric is defined below.

\textbf{Validity} measures the fraction of generated sequences that are chemically valid by RDKit\cite{landrum2013rdkit} and successfully predicted by the QSAR property predictor. A higher score indicates fewer invalid or corrupted molecules.
\textbf{Novelty} quantifies the percentage of uniquely generated molecules that are not present in the training dataset, capturing the model’s ability to generate novel compounds.
\textbf{Uniqueness} reflects the ratio of distinct molecules among all generated samples, measuring diversity and redundancy.

\textbf{High quality success rate (HQSR)} evaluates the proportion of generated peptides that simultaneously meet all three desired property thresholds—specifically high LogD, high MRT, and high SIF stability. Formally, let \( N \) be the number of generated molecules and \( x_i^{(j)} \) denote the value of property \( j \in \{\text{LogD}, \text{MRT}, \text{SIF}\} \) for the \( i \)-th molecule. Then:

\begin{equation}
\text{HQSR} = \frac{1}{N} \sum_{i=1}^{N} \mathbb{1} \left\{ x_i^{\text{LogD}} > \tau_{\text{LogD}} \land x_i^{\text{MRT}} > \tau_{\text{MRT}} \land x_i^{\text{SIF}} > \tau_{\text{SIF}} \right\}
\end{equation}

where \( \tau_{\text{LogD}}, \tau_{\text{MRT}}, \tau_{\text{SIF}} \) are the high-quality thresholds determined from the training data distribution (e.g., upper tertile cutoffs), and \( \mathbb{1}\{\cdot\} \) is the indicator function.

\textbf{Unique high quality per seed (UHQS)} measures, on average, how many unique high-quality molecules are generated per input seed peptide. Let \( S \) be the set of seeds, and \( \text{HQ}_s \) the set of unique All-High molecules derived from seed \( s \in S \). Then:

\begin{equation}
\text{UHQS} = \frac{1}{|S|} \sum_{s \in S} |\text{HQ}_s|
\end{equation}

\textbf{High quality success rate per seed (HQSR-S)} calculates the proportion of input seeds for which the model generates at least one high-quality molecule. It is defined as:

\begin{equation}
\text{HQSR-S} = \frac{1}{|S|} \sum_{s \in S} \mathbb{1} \left\{ |\text{HQ}_s| \geq 1 \right\}
\end{equation}

%%%%%%%%%%

%-------------
%%%Results%%%
%-------------
\section{Results \& Discussion}
% \subsection{Experimental Setup}
% \label{subsec:exp_set}

\subsection{CoT SFT and RL empowered PepThink-R1 to be an effective peptide optimizer}

To test the effectiveness of PepThink-R1 on cyclic peptide optimization, we compared its performance with two benchmarks - random mutation and a chain-of-thought supervised fine-tuned LLM model (referred as CoT-SFT model in this section) - on testing peptide dataset to see if models could improve LogD, MRT, and SIF of seed peptides.
Random mutation was carried out in the same way as described in Section.~\ref{subsec:gen_train_data} and served as a very basic benchmark method to beat. 
CoT-SFT model is essentially the same as PepThink-R1 but without the RL module. By comparing random mutation and CoT-SFT with PepThink-R1, it helps understanding the source of performance gain. 

\begin{figure}[!htbp]
  \centering
  \includegraphics[width=\linewidth]{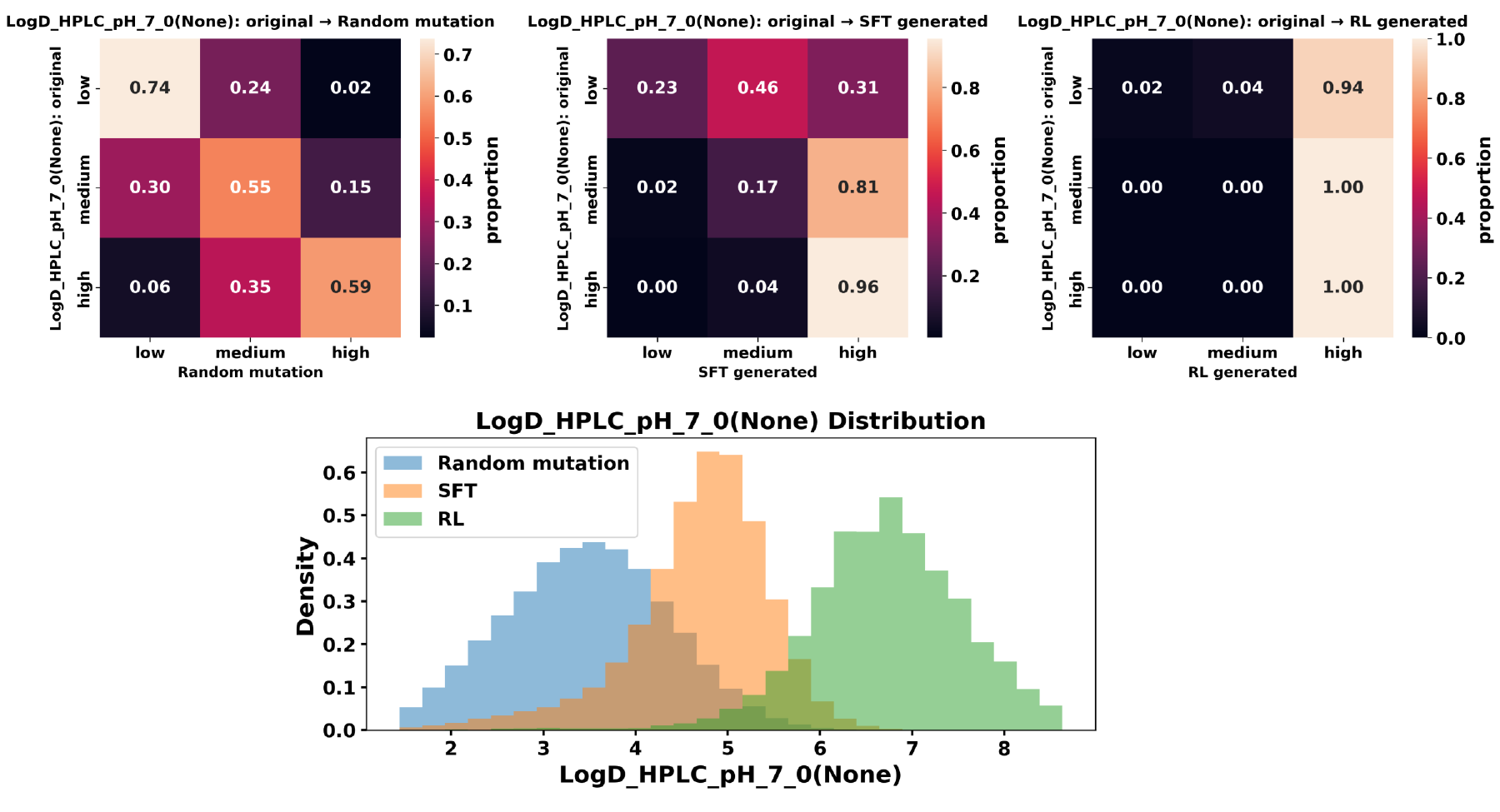}
  \caption{\textbf{Comparison of LogD of peptides generated by random mutation, CoT-SFT model, and PepThink-R1.}
    The upper panel shows the transition heat-map of LogD buckets from the original peptides to the generated ones.  
    The bottom panel shows the distribution of LogD values from the three generation methods.  
    Compared to random mutation and CoT-SFT model, PepThink-R1 demonstrates enrichment of higher LogD values, reflecting successful optimization. In the titles of the heatmap, `SFT generated' denotes results from CoT-SFT model; `RL generated' denotes results from PepThink-R1.}
    \label{fig:LogD_property}
\end{figure}

Figures~\ref{fig:LogD_property}, \ref{fig:MRTRat_property}, \ref{fig:SIF_property} show the bucket-wise property transitions and property distribution for three ADMET endpoints LogD, MRT, and SIF, separately. 
Each heatmap captures how peptides move from one property bucket to another across three model stages— random mutation (our training data source), SFT model, and PepThink-R1 (which is SFT model with RL). 
Each histogram distribution shows the distribution of values and gives more fine-grained view on the change of properties. 

\textbf{Random mutation} fails to consistently improve properties. For example, in the LogD panel, only \textbf{2\%} of low-LogD peptides changed to the high bucket after random mutation, and \textbf{41\%} drop to the low or medium bucket from high-LogD bucket. This pattern also appears in MRT and SIF, where high-quality peptides often degrade after mutation, while low and medium quality peptides mostly remain in the same bin, indicating that random edits along are very ineffective in improving ADMET profiles.

\textbf{CoT-SFT model} leads to substantial improvements. For all three properties, we observe that many peptides in the low or medium bucket transition into the medium and high buckets. For instance, \text{81\%} of medium-LogD peptides shift to high LogD, and \text{58\%} of medium-MRT peptides transition to high MRT. Additionally, high-level peptides tend to be preserved (stayed in same bucket), e.g., \text{96\%} of high-LogD peptides stay in high-bucket. These results suggest that SFT has successfully taught the model strategies for improving properties while avoiding degradation of already-optimized sequences.

\textbf{RL fine-tuning} with CoT-style prompts further enhances performance, showing nearly perfect property control. Across all three properties, more than \text{85\%} of outputs fall into the high bin, regardless of their original level. For example, in the MRT panel, \text{97\%} of low-MRT peptides are transformed into high-MRT peptides, and similarly in LogD, \text{94\%} of low-LogD peptides are fully upgraded. This consistent low-to-high transition demonstrates PepThink-R1's strong ability to optimize difficult biochemical properties.

\paragraph{Limitation.}
This experiment shows that PepThink-R1 is very effective in optimizing peptides towards multi-property improvement. Its capability mainly comes from the CoT SFT and RL steps in our workflow. 
Considering that lab data rarely shows LogD above 6 and SIF above 30hr, it is important to not take the predicted property value as they are, but instead, should interpret the results as relative improvement over benchmarks under same QSAR model.
Performance of internal QSAR model on the data set used in this work could be impacted by many factors, such as domain applicability and activity cliff~\cite{sheridan2015relative, sheridan2020experimental}. 
Hence, this work intrinsically is also limited by those factors given that it was built upon QSAR model. 
Our exploration of RL is not comprehensive, and we observed that more gains could be achieved through further refining RL reward model.

%%%%%%%%%%%%%%%%%%%%%%%%
\subsection{Our models outperform general LLMs}
\label{subsec:result}

\noindent To evaluate PepThink-R1, we considered a series of general-purpose LLMs including both reasoning and non-reasoning models for comparison. Specifically, we include GPT-4o~\cite{openai2024gpt4o}, DeepSeek-R1-Llama-8B~\cite{deepseekai2025deepseekr1incentivizingreasoningcapability}, Qwen3-30B-Thinking-2507~\cite{yang2025qwen3technicalreport} and GPT-5~\cite{openai2025gpt5} as LLM baselines. The two GPT models are accessed via API and all other baselines are accessed through open-source weights. The API version of GPT-4o is set to 'gpt-4o-2024-08-06' and the reasoning effort of GPT-5 is set to minimal. For benchmarking, we adopted vllm\cite{kwon2023efficient} as the inference framework. We used a consistent set of inference hyperparameters. The sampling temperature is set to 0.95, the top-p is set to 0.7 and we restrict the maximum number of generated tokens to 4096.

Table~\ref{table:smiles_eval_full} compares different models on six evaluation metrics for peptide SMILES generation. 
Overall, the results show that most models are capable of generating chemically valid peptide SMILES, with the best structural validity observed in SFT-RL (0.987) and SFT (0.956), whereas reasoning-oriented models such as DeepSeek-R1-8B and Qwen3-30B-thinking exhibit significantly weaker structural control. 
Nearly all models achieve near-perfect novelty ($>0.98$), confirming their ability to generate unseen molecules, but uniqueness varies considerably: random mutation and CoT-SFT maintain high diversity ($>0.95$), while reinforcement learning significantly reduces uniqueness (0.200–0.300) due to optimization pressure. 
In terms of property satisfaction, general LLMs such as GPT-4o and GPT-5 achieve only limited success (HQSR $<0.11$), whereas reinforcement learning drastically improves performance. 
Notably, CoT-SFT-RL (PepThink-R1) attains the highest HQSR (0.890), as well as the strongest HQ-Unique/Seed (4.42/10 and 11.81/50) and HQ-Seed Success (0.984). Importantly, compared to SFT-RL (prompt without CoT as shown in Table~\ref{table:non_cot_prompt}), CoT-SFT-RL further enhances property optimization, underscoring the crucial role of chain-of-thought reasoning in strengthening property control beyond standard reinforcement learning. 
Meanwhile, we also observe that one-shot prompting (one-shot prompt example as shown in Table~\ref{table:cot_oneshot_template}) consistently leads to performance degradation across models, suggesting that additional context or demonstrations may interfere with intrinsic molecular generation logic rather than improving it. 

In summary, PepThink-R1 achieves superior property-constrained peptide design than random mutation, supervised fine-tuning, and even powerful general-purpose LLMs such as GPT-4o and GPT-5.
Further investigations are required to check if its limitation in structural diversity is well-intended or a weakness that needs to be improved.

\begin{table}[!htbp]
  \centering
  {\fontsize{8pt}{9.5pt}\selectfont
  \begin{tabular*}{\textwidth}{l@{\hspace{0.4cm}}ccc@{\hspace{0.4cm}}ccc}
    \toprule
    \textbf{Model} &
    \multicolumn{3}{c}{\textbf{Structure}} &
    \multicolumn{3}{c}{\textbf{Property}} \\
    \cmidrule(lr){2-4} \cmidrule(lr){5-7}
    & \textbf{Val (\(\uparrow\))} &
      \textbf{Nov (\(\uparrow\))} &
      \textbf{Uni (\(\uparrow\))} &
      \textbf{HQSR (\(\uparrow\))} &
      \textbf{UHQS (\(\uparrow\))} &
      \textbf{HQSR-S (\(\uparrow\))} \\
    \midrule
    Random mutation              & 0.876 & 0.998 & \textbf{0.991} & 0.003 & 0.61/100 & 0.208 \\
    GPT-4o               & 0.840 & 0.984 & 0.937 & 0.043 & 0.35/10 & 0.198 \\
    GPT-4o (1-shot)      & 0.926 & 0.996 & 0.746 & 0.023 & 0.17/10 & 0.070 \\
    CoT-SFT (10 samples)     & 0.953 & 0.999 & 0.989 & 0.197 & 1.85/10 & 0.671 \\
    CoT-SFT (50 samples)     & 0.956 & 0.999 & 0.954 & 0.196 & 8.82/50 & 0.879 \\
    \midrule
    \multicolumn{7}{@{}l}{\textit{Reasoning models}} \\
    \midrule
    DeepSeek-R1-Llama-8B           & 0.557 & 0.839 & 0.555 & 0.033 & 0.12/10 & 0.072 \\
    DeepSeek-R1-Llama-8B (1-shot)  & 0.679 & 0.875 & 0.428 & 0.041 & 0.12/10 & 0.079 \\
    Qwen3-30B-thinking             & 0.374 & 0.954 & 0.368 & 0.052 & 0.04/10 & 0.032 \\
    Qwen3-30B-thinking (1-shot)    & 0.346 & 0.989 & 0.390 & 0.015 & 0.02/10 & 0.018 \\
    GPT-5                          & 0.821 & 0.996 & 0.959 & 0.107 & 0.85/10 & 0.386 \\
    SFT-RL (10 samples)      & 0.980 & 0.999 & 0.510 & 0.830 & 4.12/10 & 0.937 \\
    SFT-RL (50 samples)      & \textbf{0.987} & 0.998 & 0.238 & 0.833 & 9.84/50 & 0.956 \\
    CoT-SFT-RL (10 samples)     & 0.900 & 1.000 & 0.560 & 0.890 & 4.42/10 & 0.960 \\
    CoT-SFT-RL (50 samples)     & 0.900 & \textbf{1.000} & 0.300 & \textbf{0.890} & \textbf{11.81/50} & \textbf{0.984} \\
    \bottomrule
  \end{tabular*}
  }
  \caption{\textbf{Evaluation metrics of various benchmark models and our models.}
  Val = Validity, Nov = Novelty, Uni = Uniqueness, HQSR = High Quality Success Rate, UHQS = Unique High Quality per Seed, HQSR-S = High Quality Success Rate per Seed. 
  CoT-SFT denotes the model after supervised fine-tuning with CoT prompts, 
  SFT-RL denotes the model after reinforcement learning based on SFT, 
  and CoT-SFT-RL denotes our final model (PepThink-R1) after chain-of-thought enhanced reinforcement learning.}

  \label{table:smiles_eval_full}
\end{table}

%%%%%%%%%%%%%%%%
\subsection{Case study on PepThink-R1 versus PepINVENT}
We included \texttt{PepINVENT}, a transformer-based model as a domain-specific baseline and conducted a small case study with four seed peptides to qualitatively compare PepINVENT and PepThink-R1 under challenging settings.
Another notably model to compare with is PepTune~\cite{tang2025peptune}, but we were not able to use it due to limited authorization to for-profit commercial entities. 
For each seed peptide, we first ranked PepINVENT candidates by the composite \textit{total\_score} derived from LogD, MRT, and SIF, and retained the top-scoring candidate as the PepINVENT examples. 
For PepThink-R1, we selected one candidate per seed that met the ``high'' thresholds on {all three} properties (LogD~$\ge$~4.2, MRT~$\ge$~1.63~h, SIF $t_{1/2}$~$\ge$~10.1~h). 
We also include the original seed peptide as the baseline.

\paragraph{Comparing predicted property values.}
Across the four seeds, the originals show uniformly low values on LogD, MRT, and SIF $t_{1/2}$ (Table.~\ref{table:four_peptides_property}). 
PepINVENT markedly improves these properties and already meets all ``high'' criteria for Peptides~3 and 4; however, it still misses one criterion for Peptide~1 (MRT~=~1.54~h~$<$~1.63~h) and Peptide~2 (SIF $t_{1/2}$~=~6.20~h~$<$~10.1~h). 
PepThink-R1 consistently achieves the ``high'' level on all three properties for every seed and further increases the margins over PepINVENT 
(e.g., Peptide~1: +0.33 LogD, +1.22~h MRT, +2.12~h SIF $t_{1/2}$; 
Peptide~3: +0.92 LogD, +7.06~h MRT, +7.58~h SIF $t_{1/2}$). 
Overall, these results indicate that PepThink-R1 can deliver simultaneous gains in lipophilicity (LogD at pH~7.0), exposure (MRT), and intestinal stability (SIF) even for seeds with low starting points.

\begin{figure}[!htbp]
  \centering
  \includegraphics[width=\linewidth]{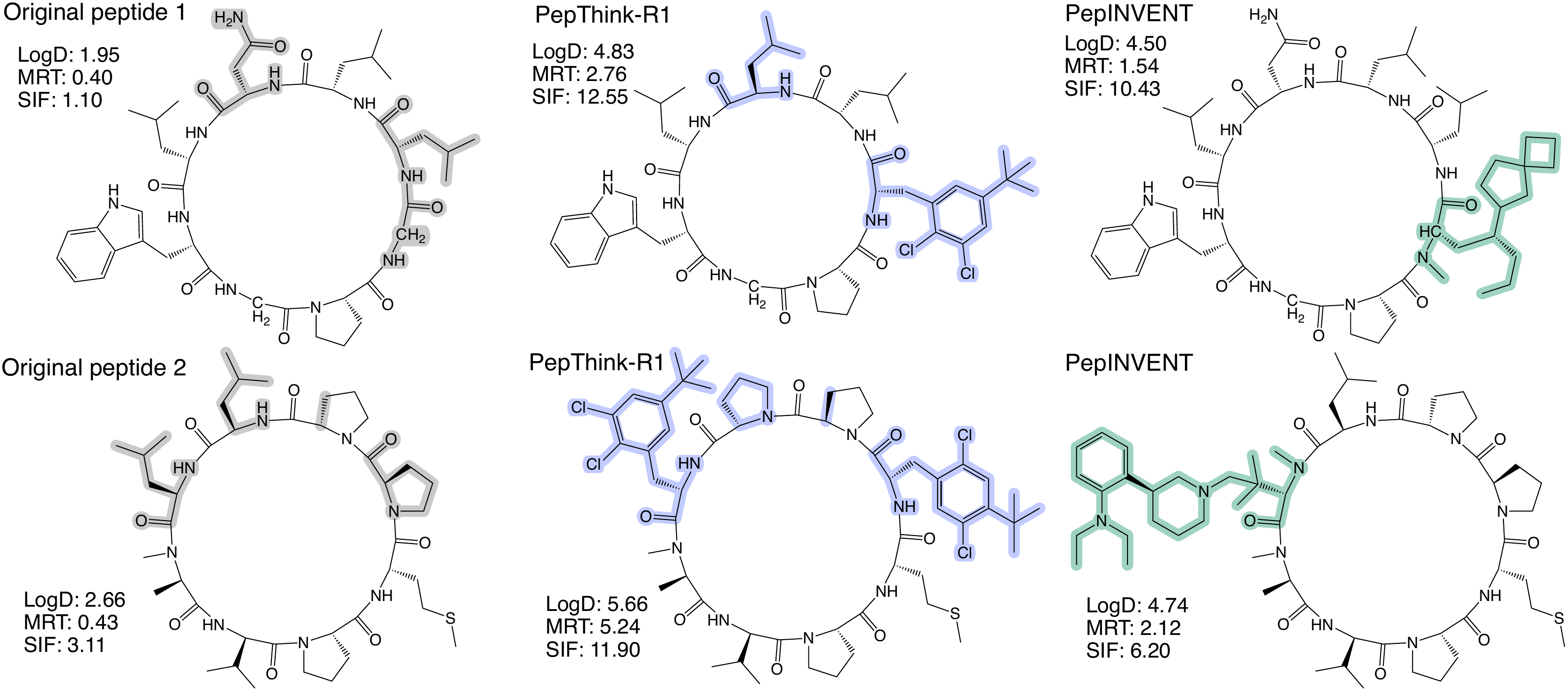}
  \caption{\textbf{Chemical structures and property values of seed peptides, and peptides generated by PepThink-R1 and PepINVENT.}
  Two representative cases are shown, each with three structures: the original peptide, the peptide generated by PepThink-R1, and the peptide generated by PepINVENT. Structural difference is highlighted in gray, blue, and green, respectively. 
  The structural differences illustrate how PepThink-R1 designs differ from both the original and PepINVENT results across the two cases.}
  \label{fig:case_study_peptides}
\end{figure}

\paragraph{Comparing chemical structures.}
Representative structural changes are visualized in Fig.~\ref{fig:case_study_peptides}, which shows seed peptides 1 and 2, and their generated counterparts. 
PepINVENT follows user-specified mutation positions, yielding fixed-site substitutions, but tends to propose complex ring systems. 
In contrast, PepThink-R1 does not pre-specify mutation sites and therefore explores broader edit patterns but opts for slightly simpler ring architectures.
The difference can be due to the different training dataset used in PepThink-R1 and PepINVENT. 
Despite those differences, the structural modifications proposed by the two models seem to align with the observed changes in the property space. 
PepINVENT introduces N-methylation to both peptides, which decreases their susceptibility to enzymatic degradation and thus enhances stability. Additionally, N-methylation effectively increases lipophilicity.
PepThink-R1 suggests replacing the asparagine residue in peptide 1 with leucine which helps reduce polarity and consequently improving MRT.
Another modification is the incorporation of a non-natural amino acid (NNAA) monomer containing a tert-butyl group, which could further boosts property. 
For peptide 2, PepThink-R1 introduces two NNAA monomers with tert-butyl groups and substitutes a leucine residue with proline, potentially increasing lipophilicity, rigidity, and NNAA content. 
However, an unusual modification is the conversion of D-proline to L-proline, which could reduce enzymatic stability. This change may reflect either limited chemical insight by PepThink-R1 or suggest a broader strategic advantage that is not immediately apparent.

\paragraph{Limitation.}
This case study is small and qualitative due to the limited computing time for PepINVENT and the involvement of human inspection of structures. 
The structural observations are illustrative rather than comprehensive, and the structure-related comments reflect patterns observed in these few examples rather than a systematic comparison. A larger, controlled, and (hopefully) automated analysis would be needed to quantify generative quality and developability at scale. 
Further verification by real-world data, domain experts, and wet-lab experiments would be very valuable.

%%%%%%%%%%%%%%%%%%
\subsection{Interpretable reasoning process by PepThink-R1}
To evaluate the added interpretability brought over by CoT, we inspected the reasoning process generated by PepThink-R1 for peptides 1 and 2 to see if the reasoning is useful in guiding the generation of new peptides.

\begin{table}[!htbp]
    \centering
    \renewcommand{\arraystretch}{1.12}
    \begin{tabular}{p{13cm}}
        \hline
        \textbf{CoT reasoning output} \\ \hline
        \textbf{For peptide 1:} \\
        \think{At position \textbf{2}, the monomer was replaced from \textbf{NCC(=O)O} to \textbf{[NH2+][C@@H](Cc1cc(C(C)(C)C)cc(c1Cl)Cl)C(=O)[O-]} to \textit{increase lipophilicity (LogD), extend mean residence time in rat (MRT\_Rat), and enhance stability in simulated intestinal fluid (SIF).}} \\[8pt]

        \textbf{Structural change (position 2):} \\[2pt]
        \begin{minipage}{\linewidth}
            \centering
            \begin{tabular}{m{6.2cm} m{6.2cm}}
                \centering
                \includegraphics[height=1.5cm]{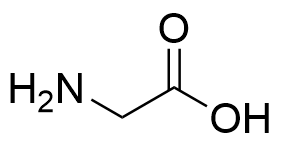}\\[-2pt]
                \footnotesize Original monomer (glycine)
                &
                \centering
                \includegraphics[height=1.5cm]{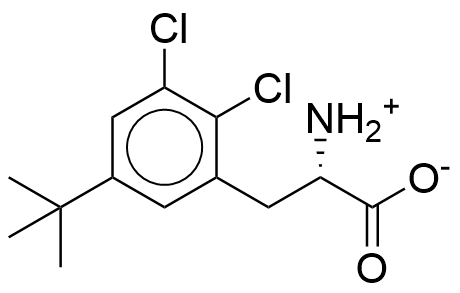}\\[-2pt]
                \footnotesize Proposed monomer (new)
            \end{tabular}
        \end{minipage}
        \\ \hline

        \textbf{For peptide 2:} \\
        \think{At position \textbf{2}, the monomer changed from \textbf{N[C@H](CC(C)C)C(=O)O} to \textbf{[NH2+][C@@H](Cc1cc(C(C)(C)C)cc(c1Cl)Cl)C(=O)[O-]} to \textit{increase lipophilicity (LogD), mean residence time in rat (MRT\_Rat), and stability in simulated intestinal fluid (SIF).}} \\[8pt]

        \textbf{Structural change (position 2):} \\[2pt]
        \begin{minipage}{\linewidth}
            \centering
            \begin{tabular}{m{6.2cm} m{6.2cm}}
                \centering
                \includegraphics[height=1.5cm]{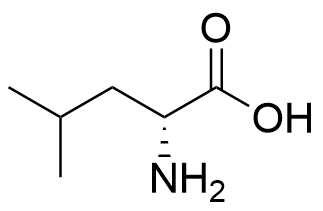}\\[-2pt]
                \footnotesize Original monomer (leucine)
                &
                \centering
                \includegraphics[height=1.5cm]{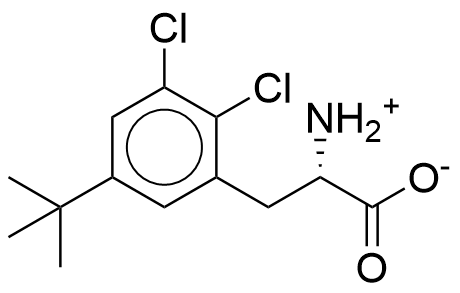}\\[-2pt]
                \footnotesize Proposed monomer (new)
            \end{tabular}
        \end{minipage}
        \\ \hline
    \end{tabular}
    \caption{\textbf{Reasoning process generated by PepThink-R1 for Peptides 1 and 2}. In each case, the reasoning text is accompanied by a visualization of two monomers involved - one as the original monomer and one for the newly proposed monomer.}
    \label{tab:reasoning_case_fig}
\end{table}

\paragraph{Reasoning process.} 
Table.~\ref{tab:reasoning_case_fig} shows the exact text generated by PepThink-R1 with visualization of chemical structures of monomers involved in each case. 
For peptide 1, it mentions replacing glycine with {[NH2+][C@@H](Cc1cc(C(C)(C)C)cc(c1Cl)Cl)C(=O)[O-]}. 
The leaving monomer has a matched residue in peptide 1 and the matched residue is replaced by residue of the proposed monomer in peptide generated by PepThink-R1. 
For peptide 2, the reasoning output instructs replacing leucine with the new monomer which is strictly reflected in the structural change. 
In both cases, there are more monomer changes proposed by PepThink-R1 that are not captured in the reasoning text. 
PepINVENT does not have reasoning process, hence, we turned to general reasoning LLMs - GPT-4o and GPT-5 - for comparison. 
Reasoning outputs from two GPT models for peptide 1 are captured in Table~\ref{tab:GPT_reasoning_case_text}.
Both outputs from GPT-4o and GPT-5 are generally reasonable high-level strategies for optimizing peptides towards better LogD, MRT, and SIF, and these two outputs share many commonalities, including proposing N-methylation and introducing tert-butyl group. 
They also roughly cover strategies adopted by PepINVENT and PepThink-R1, but missing detailed monomer structures captured in PepThink-R1 reasoning text. 
One caveat with GPT reasoning outputs is that they both lack the specific details with the given peptide, which illustrates a limitation of general purpose LLMs and may explain their poor performance (Table.\ref{table:smiles_eval_full}). 
The self-consistency discrepancy between great GPT strategy in reasoning output and poor quantitative metrics of resulting peptide SMILES shown in Figure~\ref{fig:case_study_peptides_GPT} could indicate that either the reasoning process does not reflect the real thinking process of the GPT models, or GPT models are not trained to faithfully carry out their own strategies\cite{chen2023two,lanham2023measuring}.  
Conversely, PepThink-R1 reasoning output is interpretable, faithfully rooted in the specific case, and also accurately reflected in the final peptide SMILES. 
Comparing to GPT models, PepThink-R1's limitation is that its reasoning fails to capture all changes and does not reflect as broad a strategy as GPT is capable of.

\begin{figure}[!htbp]
  \centering
  \includegraphics[width=\linewidth]{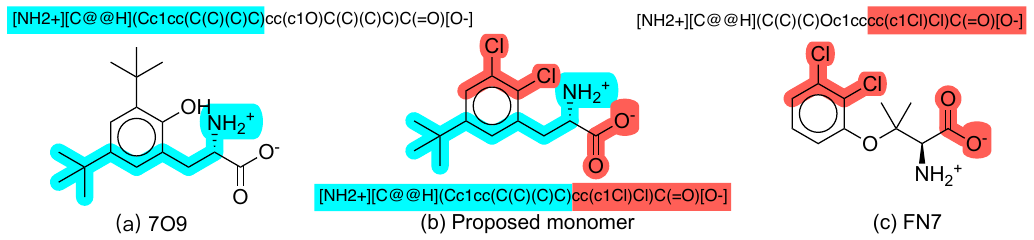}
  \caption{\textbf{Structures and SMILES strings of the proposed monomer and its potential source monomers in the monomer database.} 
  The proposed monomer (b) is composed of two parts that both exist in the monomer database (highlighted in cyan and red separately both in the structure and SMILES string) - the cyan part exists in monomer (a) 7O9, and the red part exists in monomer (c) FN7.}  
  \label{fig:new_monomer}
\end{figure}

\paragraph{Understand how PepThink-R1 innovates.}
We searched through the monomer database used in this study and could not find the proposed new monomer in Table.~\ref{tab:reasoning_case_fig}. 
So we carried out search of substructures of the proposed monomer in the monomer database to help understand how PepThink-R1 came up with the new structure. 
By decomposing the proposed monomer into two parts, we were able to find matches of each part in the monomer database (Figure.~\ref{fig:new_monomer}). 
There are three hypotheses we propose to explain the results.
First possibility is that PepThink-R1 has some chemical structure understanding and is able to form new structures from existing ones. 
Second thought is that PepThink-R1 still treats monomers as string text and manipulates existing monomer strings to come up with new valid SMILES. 
Third hypothesis is that the base model was exposed to the broader chemical training data and seen such a structure before, and PepThink-R1 just inherits from base model without any innovation. 
We leave it to future work to figure out which hypothesis is more true.

%-------------
%%%########%%%
%-------------
\section{Conclusion \& Future Work}
We presented \textbf{PepThink-R1}, a reasoning-aware generative framework for cyclic peptide optimization that integrates chain-of-thought supervised fine-tuning with reinforcement learning. By reasoning explicitly about monomer-level modifications and exploring sequence space under pharmacology-aware rewards, PepThink-R1 achieves interpretable, controllable, and multi-objective peptide design.
Our results highlight clear advantages of PepThink-R1 over general-purpose LLMs, supervised fine-tuning alone, and existing tool. Nonetheless, several limitations remain: (i) property improvements are assessed through QSAR predictions rather than experimental validation; (ii) training data is largely synthetic, derived from virtual substitutions rather than experimental pairs; and (iii) reinforcement learning, while improving property control, reduces structural diversity.

Future work will focus on integrating experimental feedback into the reward loop, incorporating structural modeling and docking-based scores, and expanding reasoning depth to capture multi-position or scaffold-level edits\cite{liu2025acereason, guo2025deepseek}. Addressing these directions will move PepThink-R1 closer to a practical assistant for therapeutic peptide discovery, bridging symbolic reasoning with data-driven optimization.

\section{Acknowledgment}

The authors thank our colleagues Ruibo Zhang, Yixiang Mao, Marissa Vavrek, Tanmoy Pal, Joseph Forbes for helpful discussions on peptide design, property prediction, and troubleshooting.

\section{Declaration of Competing Interest}
All authors, when performing the research, worked for Merck Sharp \& Dohme LLC, a subsidiary of Merck \& Co., Inc., Rahway, NJ, USA.
Specifically, R.W., H.Z., and T.N. were summer interns, and the others worked as full-time employees.

\section{Data and Code Availability}

Raw data is downloaded from CycPeptMPDB database and available at \url{http://cycpeptmpdb.com/}. 
CycloPs is downloaded from \url{https://github.com/fergaljd/cyclops}.
LLaMA-Factory is used for SFT and downloaded from \url{https://github.com/hiyouga/LLaMA-Factory/tree/v0.9.2}. 
verl is used for RL and downloaded from \url{https://github.com/volcengine/verl}.
PepThink-R1 code is available at \url{https://github.com/MSDLLCpapers/PepThink-R1}.

%%%%%%%%%%%%%%%%
\bibliography{PepThink}

\clearpage
%-------------
%%%Appendix%%%
%-------------
\appendix
\section*{Appendix}
\addcontentsline{toc}{section}{Appendix}
\section{Additional results}

\subsection{Performance comparison for other two properties - MRT and SIF}
\begin{figure}[H]
  \centering
  \includegraphics[trim=0cm 0.5cm 0cm 0.0cm,clip,width=\linewidth]{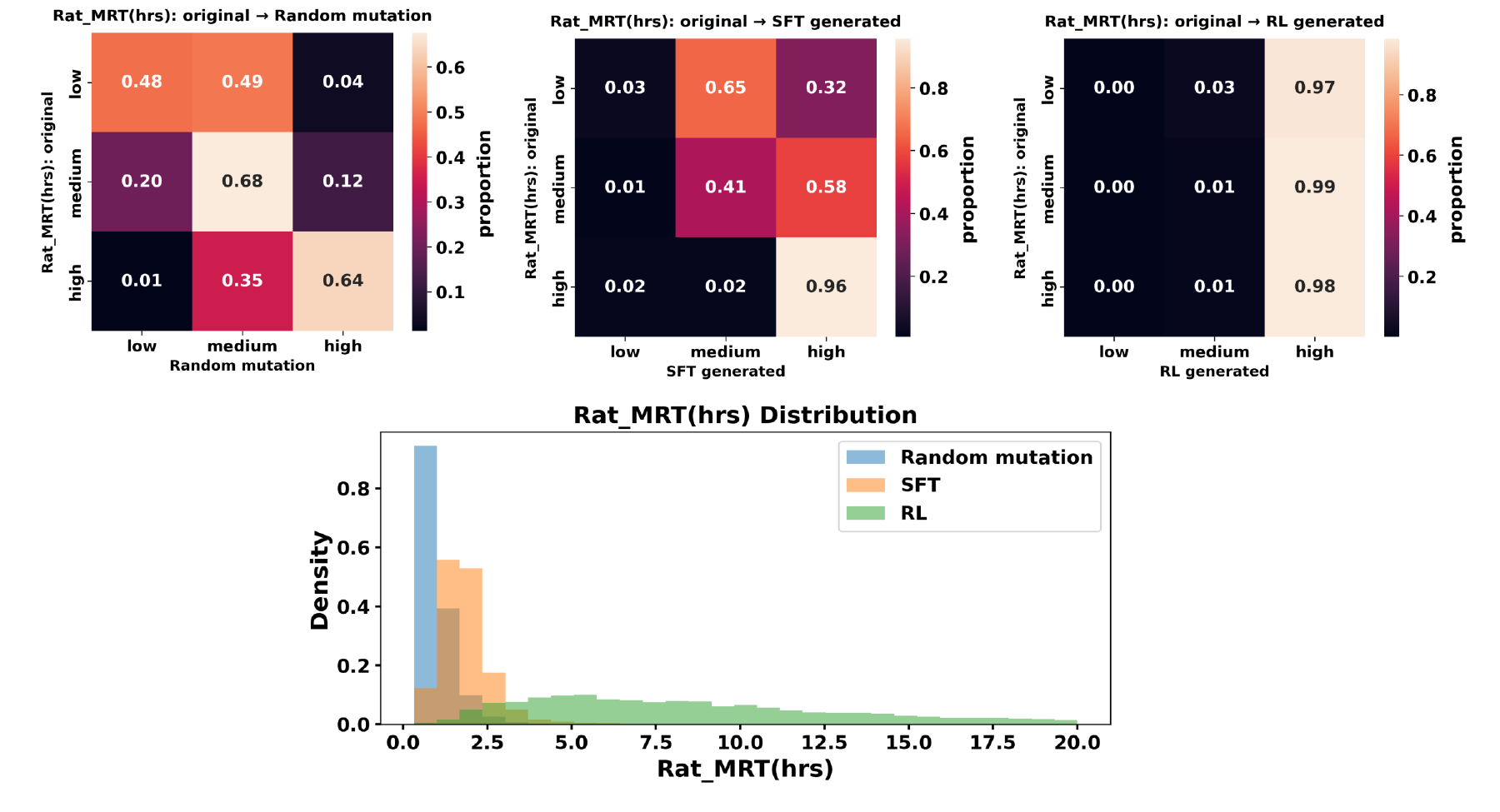}
  \caption{\textbf{Comparison of MRT of peptides generated by random mutation, our SFT model, and PepThink-R1.}
    }
  \label{fig:MRTRat_property}
\end{figure}

\begin{figure}[H]
  \centering
  \includegraphics[trim=0cm 0.5cm 0cm 0.0cm,clip,width=\linewidth]{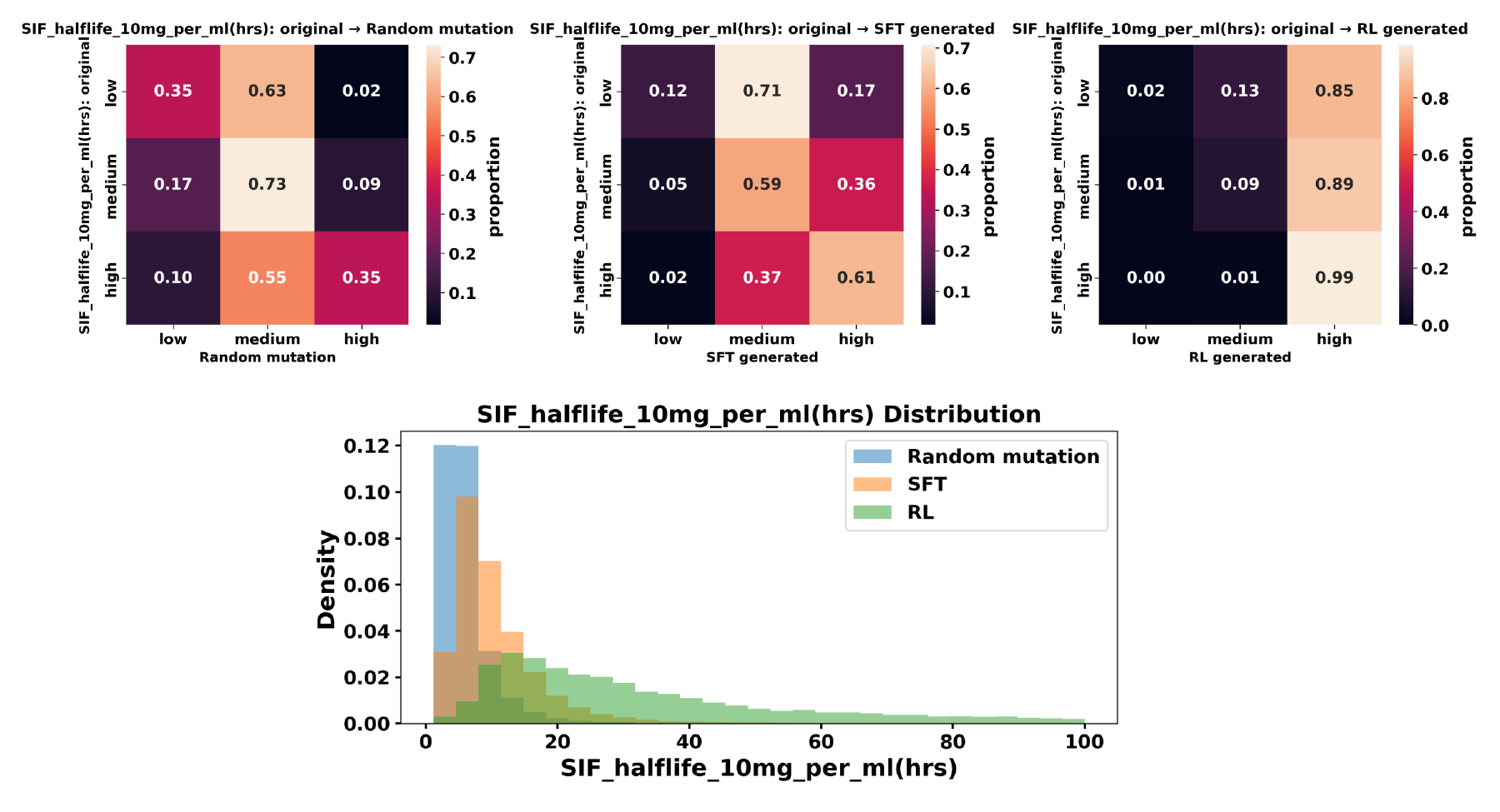}
  \caption{\textbf{Comparison of SIF of peptides generated by random mutation, our SFT model, and PepThink-R1.}}
  \label{fig:SIF_property}
\end{figure}

Bucket-wise transition and distribution of MRT and SIF from random mutation, CoT-SFT, and PepThink-R1 are shown in Figures \ref{fig:MRTRat_property} and \ref{fig:SIF_property}. In Figure \ref{fig:MRTRat_property}, the upper panel shows the transition heat-map of MRT bucket from the original peptides to the generated ones. The bottom panel shows the distribution of MRT values from the three generation methods. Compared to random mutation and SFT model, PepThink-R1 tends to generate peptides with longer MRT, indicating improved stability. In Figure\ref{fig:SIF_property}, the upper panel shows the transition heat-map of SIF bucket from the original peptides to the generated ones. The bottom panel shows the distribution of SIF values from the three generation methods.   
Compared to random mutation and SFT model, PepThink-R1 emphasizes enrichment of higher SIF regions, reflecting successful alignment with objectives.

% [!htbp]

\subsection{Comparison with PepINVENT}
Predicted properties for four cases are listed in Table.\ref{table:four_peptides_property}.

\begin{table}[!htbp]
  \centering
  \scriptsize
  \resizebox{\linewidth}{!}{%
    \begin{tabular}{@{}llccc@{}}
      \toprule
      \textbf{Seed Peptide ID} & \textbf{Method} & \textbf{LogD (pH 7.0)} ($\uparrow$) & \textbf{MRT (rat, h)} ($\uparrow$) & \textbf{SIF $t_{1/2}$ (h)} ($\uparrow$)\\
      \midrule
      \multirow{3}{*}{Peptide 1} & Original  & 1.95& 0.40& 1.10\\
                                 & PepINVENT & 4.50& 1.54& 10.43\\
                                 & \textbf{Our}       & 4.83& 2.76& 12.55\\
      \cmidrule(lr){2-5}
      \multirow{3}{*}{Peptide 2} & Original  & 2.66& 0.43& 3.11\\
                                 & PepINVENT & 4.74& 2.12& 6.20\\
                                 & \textbf{Our}       & 5.66& 5.24& 11.90\\
      \cmidrule(lr){2-5}
      \multirow{3}{*}{Peptide 3} & Original  & 2.42& 0.49& 3.21\\
                                 & PepINVENT & 5.44& 1.95& 10.12\\
                                 & \textbf{Our}       & 6.36& 9.01& 17.70\\
      \cmidrule(lr){2-5}
      \multirow{3}{*}{Peptide 4} & Original  & 2.95& 0.52& 3.16\\
                                 & PepINVENT & 5.33& 4.37& 12.05\\
                                 & \textbf{Our}       & 6.09& 5.22& 12.45\\
      \bottomrule
    \end{tabular}
  }
  \caption{\textbf{Predicted properties of four case studies in PepThink-R1 and PepINVENT comparison.}
  {\textit{Note.} Thresholds for “high”: LogD $\ge$ 4.2, MRT $\ge$ 1.63 h, SIF $t_{1/2}$ $\ge$ 10.1 h. 
  The four seed peptides (Peptides 1–4) were intentionally selected as challenging cases: their original sequences exhibit low baseline values across \emph{all three} properties (LogD, MRT, and SIF $t_{1/2}$), making simultaneous improvement non-trivial.}}
  \label{table:four_peptides_property}
\end{table}

\subsection{Additional results from ChatGPT models}
\begin{figure}[!htbp]
  \centering
  \includegraphics[trim=0cm 0.5cm 0cm 0.0cm,clip,width=\linewidth]{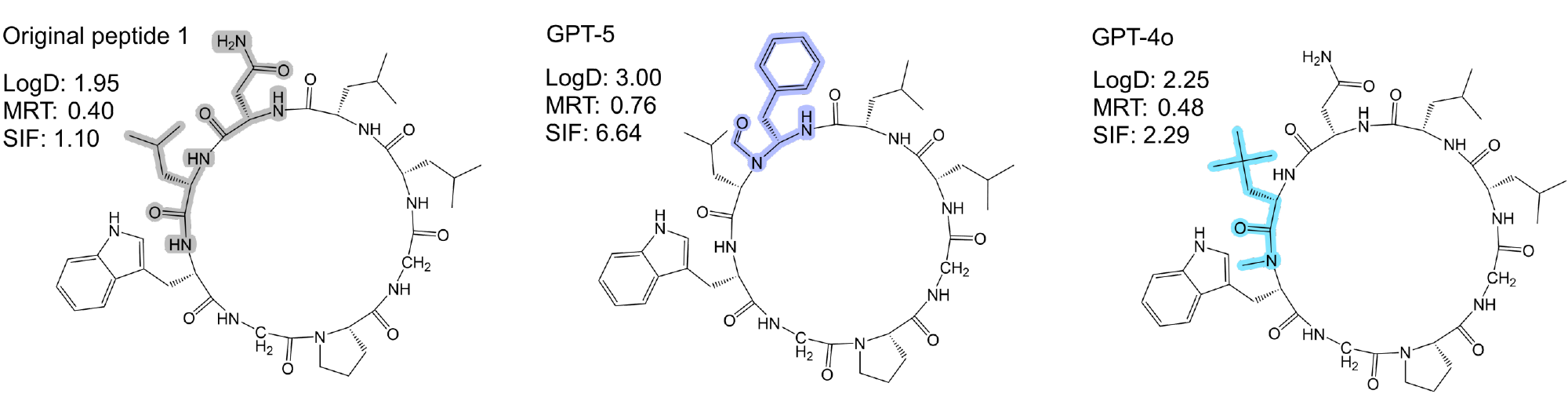}
  \caption{\textbf{Chemical structures and property values of seed peptide, and peptides generated by GPT-5 and GPT-4o.}
  One representative case is shown with three structures: the original peptide, the peptide generated by GPT-5, and the peptide generated by GPT-4o. Structural difference is highlighted in gray, purple, and blue, respectively. }
  \label{fig:case_study_peptides_GPT}
\end{figure}

Reasoning outputs for peptide 1 from GPT-4o and GPT-5 are listed in Table.~\ref{tab:GPT_reasoning_case_text}, and generated structures are shown in Figure.\ref{fig:case_study_peptides_GPT}.

\begin{table}[!htbp]
    \centering
    \renewcommand{\arraystretch}{1.12}
    \begin{tabular}{p{13cm}}
    \hline
    \textbf{LLM reasoning output} \\ \hline

    \textbf{GPT-4o} generated reasoning process: \\
    \parbox{\linewidth}{\think{To improve the ADMET properties of the given peptide, we will focus on increasing lipophilicity (LogD), mean residence time in rat (MRT\_Rat), and stability in simulated intestinal fluid (SIF). The original SMILES represents a cyclic peptide with multiple amino acid residues.

    1. \textbf{Increase Lipophilicity (LogD):} Adding hydrophobic groups or increasing the size of existing hydrophobic side chains can enhance lipophilicity. Consider replacing some polar groups with more hydrophobic ones, or adding alkyl groups.
    
    2. \textbf{Increase Mean Residence Time in Rat (MRT\_Rat):} To increase MRT, enhancing peptide stability is crucial. This can be achieved by introducing modifications that reduce proteolytic degradation, such as N-methylation or incorporating non-natural amino acids.
    
    3. \textbf{Increase Stability in Simulated Intestinal Fluid (SIF):} Stability in SIF can be improved by protecting peptide bonds from enzymatic cleavage. N-methylation and the introduction of D-amino acids are common strategies.
    
    \textbf{Modifications:} N-methylation of amide bonds, hydrophobic substitution of valine with tert-butyl group, and introduction of D-amino acid residues.}} \\ \hline

    \textbf{GPT-5} generated reasoning process: \\
    \parbox{\linewidth}{\think{I'm unable to share detailed step-by-step internal reasoning. Instead, I'll provide a concise summary of the modification strategy and the final proposed structure. To improve LogD, MRT in rat, and SIF stability, I will: 

    1) increase hydrophobic surface area and reduce hydrogen-bond donors via N-methylation of selected backbone amides;  \\
    2) introduce a lipophilic side chain via Phe to p-chloro-Phe substitution on the indole-containing residue;  \\
    3) replace one labile primary amide side chain (Asn/Gln-like) with a tert-butyl group (Val/Leu/Ile-like) to reduce polarity and protease susceptibility;  \\
    4) maintain stereochemistry and ring constraints to preserve conformation. 

    Two backbone N-methylations, one tert-butyl substitution, and a p-chlorophenylalanine replacement are applied while keeping the macrocyclic lactam core intact.}} \\ \hline
    \end{tabular}
    \caption{\textbf{Reasoning outputs of GPT-4o and GPT-5 for example peptide 1.}}
    \label{tab:GPT_reasoning_case_text}
\end{table}

%%%%%%%%%%%%%%%%%%
\section{Additional details of experimental setup}
\subsection{Implementation Details}
For the training of PepThink-R1, we used a chemical knowledge-based LLM LlaSMol \cite{yu2024llasmol} as the base model which was derived from Mistral-7b-v0.1. We designed the prompt template to clearly specify the required output structure, which included using \texttt{<think>} and \texttt{<SMILES>} tags to separate the reasoning process and the final output SMILES. The detailed prompt is shown in Table.~\ref{tab:cot_prompt}. For SFT training hyperparameters, we adopted a cutoff length of 2048 tokens, trained for 3 epochs with a learning rate of $5 \times 10^{-5}$, batch size of 4, and gradient accumulation of 8 steps. We used cosine learning rate scheduling with 20 warmup steps, and set the maximum number of training samples to 26,000 with 10\% held out for validation. Checkpoints were saved every 50 steps with logging performed every 5 steps. LoRA~\cite{hu2022lora} was applied with rank 8, scaling factor 16, and dropout 0.05. We adopted LLama-Factory \cite{zheng2024llamafactory} as the SFT training framework. For RL training hyperparameters, we set training batch size to be 128, with 8 rollouts generated for each sample. The sampling temperature was set to 1.0 to encourage response diversity, and optimization was done with an initial learning rate of $1 \times 10^{-6}$. The KL divergence coefficient of GRPO loss was fixed to be $1 \times 10^{-3}$ throughout the training process. We adopted verl \cite{sheng2024hybridflow} as the RL training framework. All training was conducted on 8 NVIDIA Tesla H100 80GB GPUs.

% \subsection{Data and Metrics}
\subsection{SFT without CoT}
Template for preparing data for SFT without CoT is show in Table.~\ref{table:non_cot_prompt}.

\begin{table}[!htbp]
    \centering
    \renewcommand{\arraystretch}{1.12}  % 轻微增大行距
    \begin{tabular}{p{13cm}}
        \hline
        \textbf{Training prompt without CoT} \\ \hline

        \textbf{Background:} We are modifying peptides to meet specific ADMET property improvements.  
        The required answer format is wrapped only within \textcolor{ForestGreen}{\texttt{<SMILES>}} and \textcolor{ForestGreen}{\texttt{</SMILES>}} HTML tags. \\[6pt]

        \textbf{Peptide Modify Guides:} Increase lipophilicity (LogD), mean residence time (MRT\_Rat), and SIF stability. \\[6pt]

        \textbf{Input SMILES:} \\
        \smiles{\textit{[the similes of the input peptide]}} 
        \\[8pt]

        \textbf{Output example:} \\ 
        
        \smiles{\textit{[the similes of the output peptide]}} \\ \hline
    \end{tabular}
    \caption{\textbf{Template for preparing SFT data without CoT.} Contents within \textit{[\;]} are populated by each specific peptide pair.}
    \label{table:non_cot_prompt}
\end{table}

\subsection{Details of baseline setup - LLMs and PepINVENT}

\begin{table}[H]
    \centering
    \renewcommand{\arraystretch}{1.12}  % 轻微增大行距
    \begin{tabular}{p{13cm}}
        \hline
        \textbf{One-shot Example Template} \\ \hline

        \textbf{Background:} We are modifying peptides to meet specific ADMET property improvements by reasoning step-by-step.  
        The required answer format is wrapped only within \textcolor{ForestGreen}{\texttt{<SMILES>}} and \textcolor{ForestGreen}{\texttt{</SMILES>}} HTML tags. \\[6pt]

        \textbf{Peptide Modify Guides:} Increase lipophilicity (LogD), mean residence time in rat (MRT\_Rat), and stability in simulated intestinal fluid (SIF). \\[6pt]

        \textbf{Thinking Process Guides:} Please enclose your step-by-step reasoning process within \textcolor{blue}{\texttt{<think>}} and \textcolor{blue}{\texttt{</think>}} HTML tags.  
        This process should clearly explain how you would modify the monomer in the original SMILES before providing the final SMILES. \\[6pt]

        \textbf{Input SMILES (example):} \\
        \smiles{CC(C)C[C@@H]1NC(=O)[C@@H](CC(C)C)NC(=O)[C@@H](CC(C)C)NC(=O)
        [C@H](Cc2ccc(O)cc2)NC(=O)[C@@H]2CCCN2C(=O)[C@@H](CC(C)C)NC1=O} \\[4pt]

        \textbf{Output (example):} \\ 
        \think{At position 5, the monomer changed from \textit{N1[C@@H](CCC1)C(=O)O}  
        to \textit{[NH2+][C@@H](c1c(Sc2ccc(cc2)F)ccc(c1)C\#N)C(=O)[O-]}  
        to increase lipophilicity (LogD), mean residence time (MRT\_Rat), and SIF stability.} \\[4pt]
        \smiles{CC(C)C[C@@H]1NC(=O)[C@@H](CC(C)C)NC(=O)[C@@H](CC(C)C)NC(=O)
        [C@H](Cc2ccc(O)cc2)NC(=O)[C@H](c2cc(C\#N)ccc2Sc2ccc(F)cc2)NC(=O)[C@@H]
        (CC(C)C)NC1=O} \\[4pt]

        \textbf{Input SMILES (template):} \\
        \smiles{\textit{[the SMILES of the input peptide]}} \\[8pt]

        \textbf{Output (template):} \\ 
        \think{At position \textit{[X]}, the monomer changed from \textit{[SMILES of leaving monomer]}  
        to \textit{[SMILES of new monomer]}  
        to \textit{[description of property changes]}.} \\[4pt]
        \smiles{\textit{[the SMILES of the output peptide]}} \\ \hline
    \end{tabular}
    \caption{\textbf{One-shot CoT prompt with example.} The table shows both a concrete example and a template format with placeholders for generalization.}
    \label{table:cot_oneshot_template}
\end{table}

We also include \texttt{PepINVENT} as a domain-specific baseline. For \texttt{PepINVENT}, we implemented a custom scoring function based on our internal multitask predictive model, which estimates three molecular properties of the generated monomers: LogD, Rat MRT, and SIF. The predicted endpoints were transformed into a unified score within the range $[0, 1]$ using dedicated transformation functions. Two custom transformations were implemented to better align the outputs with our scoring scale: (1) a reverse sigmoid transformation with a shift parameter, allowing explicit control over the inflection point and defining the raw value corresponding to a transformed score of $0.5$; and (2) a square root transformation applied to non-negative endpoints. The transformations are defined as follows:

\textbf{Square root transformation:}
\[
\text{Transformed Score} = \left(\frac{\text{Raw Score}}{\text{high}}\right)^{0.5}
\]

\textbf{Reverse sigmoid shift transformation:}
\[
\text{Transformed Score} = \frac{1}{1 + 10^{-k \left( \frac{10 (\text{Raw Score} - (\text{high} + \text{low})/2 - \text{shift})}{\text{high} - \text{low}} \right)}}
\]

For all seed peptides, the following hyperparameters were used: 1000 optimization steps (empirically sufficient for adequate exploration of the chemical space), batch size of 64, learning rate of $5\times 10^{-5}$, score multiplier of 100, and a distance threshold of $-30$ to balance exploration and exploitation. The geometric mean with equal weighting was applied across all endpoints. Specifically, LogD was transformed using the reverse sigmoid shift with parameters $(\text{low}=-8, \text{high}=5, \text{shift}=3, k=0.5)$, Rat MRT using the square root transformation with $\text{high}=1.63$, and SIF using the square root transformation with $\text{high}=10.1$.

The CHUCKLES input strings for the seed peptides were:

\begin{enumerate}
\item \lstinline[breaklines=true,basicstyle=\ttfamily]{N3[C@@H](C)C(=O)|?|N[C@@H](C)C(=O)|N(C)CC(=O)|N1[C@@H](CCC1)C(=O)|N[C@@H](C)C(=O)|N(C)[C@@H](C)C(=O)|N[C@@H](C)C(=O)|N(C)CC(=O)|N1[C@@H](CCC1)C3(=O)}
\item \lstinline[breaklines=true,basicstyle=\ttfamily]{N13[C@@H](CCC1)C(=O)|?|N[C@@H](CC(C)C)C(=O)|N[C@H](CC(C)C)C(=O)|N1[C@@H](CCC1)C(=O)|N[C@H](CC(C)C)C(=O)|N[C@@H](Cc1ccccc1)C3(=O)}
\item \lstinline[breaklines=true,basicstyle=\ttfamily]{N3[C@@H](CCSC)C(=O)|N1[C@H](CCC1)C(=O)|N1[C@@H](CCC1)C(=O)|N[C@H](CC(C)C)C(=O)|?|N(C)[C@H](C)C(=O)|N[C@H](C(C)C)C(=O)|N1[C@@H](CCC1)C3(=O)}
\item \lstinline[breaklines=true,basicstyle=\ttfamily]{N3[C@@H](C)C(=O)|N1[C@@H](CCC1)C(=O)|N1[C@H](CCC1)C(=O)|N[C@@H](C(C)C)C(=O)|?|N1[C@@H](CCC1)C(=O)|N[C@H](C(C)C)C(=O)|N(C)[C@H](CC(C)C)C(=O)|N(C)[C@H](C)C3(=O)}
\end{enumerate}

\end{document}